\documentclass[runningheads]{llncs}
\usepackage{graphicx}
\usepackage{float}
\usepackage{amssymb}% http://ctan.org/pkg/amssymb
\usepackage{pifont}% http://ctan.org/pkg/pifont
\newcommand{\cmark}{\ding{51}}%
\newcommand{\xmark}{\ding{55}}%
\usepackage[linesnumbered, ruled]{algorithm2e}

\usepackage{tikz}
\usepackage{comment}
\usepackage{amsmath,amssymb} % define this before the line numbering.
\usepackage{color}
\usepackage{enumitem}
\usepackage[accsupp]{axessibility}  % Improves PDF readability for those with disabilities.

\usepackage[width=122mm,left=12mm,paperwidth=146mm,height=193mm,top=12mm,paperheight=217mm]{geometry}

\begin{document}
% \linenumbers
\pagestyle{empty}
\mainmatter
  % Insert your submission number here

\title{Task-Prior Conditional Variational Auto-Encoder for Few-Shot Image Classification} % Replace with your title

% \begin{comment}

\author{Zaiyun Yang}
\institute{$\textrm{{yzy\_dimory@stu.xjtu.edu.cn}}$}

%\end{comment}
%******************

\begin{comment}

\email{}\\
%
% \authorrunning{Y.}

\end{comment}
%******************
\maketitle

\begin{abstract}
% Few-shot learning makes a lot of significant progress in deep learning. Especially, transductive method is consistently better than inductive method. However, there is an underlying condition in transductive few-shot scenarios: the number of query shots in each class is the same. This condition may be contrary to the real world: some classes may have far less sample size than others. In this paper, to overcome this defect, we propose a Task-Prior Conditional Variational Auto-Encoder (VAE) model named TP-VAE, which is conditional on support shots and includes a task-level prior knowledge. Our model achieves state-of-the-art on a wide range of few-shot image classification tasks. Moreover, our method is a meta testing module that can connect any advanced meta training feature extractor.
% Transductive methods often outperform inductive methods in few-shot image classification scenarios. However,
Transductive methods always outperform inductive methods in few-shot image classification scenarios. However, the existing few-shot methods contain a latent condition: the number of samples in each class is the same, which may be unrealistic. To cope with those cases where the query shots of each class are nonuniform (i.e. nonuniform few-shot learning), we propose a Task-Prior Conditional Variational Auto-Encoder model named TP-VAE, conditioned on support shots and constrained by a task-level prior regularization. Our method obtains high performance in the more challenging nonuniform few-shot scenarios. Moreover, our method outperforms the state-of-the-art in a wide range of standard few-shot image classification scenarios. Among them, the accuracy of 1-shot increased by about 3\%.

% The existing transductive few-shot methods contain a latent condition:  the number of query shots in each class is the same, which may be unrealistic. Corresponding to the situation where the number of samples in some categories is different in reality, we propose a more challenging few-shot scenario ( Nonuniform few-shot learning) and we  

% Nonuniform few-shot learning, and we
% We propose a Task-Prior Conditional Variational Auto-Encoder model named TP-VAE, conditioned on support shots and constrained by a task-level prior regularization. Our method outperforms state-of-the-art on a wide range of common few-shot image classification scenarios. Among them, the accuracy of 1-shot increased by about 3\%. Moreover, we propose a more challenging few-shot scenario in which query shots in each class are different (i.e., nonuniform few-shot learning).  We then confirm that our method can maintain high performance in this scenario.

\end{abstract}

\section{Introduction}

Deep learning achieves major success in visual classification. Especially the transformer model \cite{dosovitskiy2020image,liu2021swin} and the paradigm of contrastive learning \cite{chen2020simple,he2020momentum}  set off a wave in computer vision field. However, these models and paradigms often require large amounts of data, which are unavailable in some image classification scenarios. A human can easily learn a new task rapidly with a handful of examples compared with a machine. To bridge the gap between machine and human, a new paradigm few-shot learning \cite{miller2000learning,fei2006one,vinyals2016matching}  emerges. In few-shot settings, a model is trained on the base-training set with base classes, and then we generate many few-shot tasks from the testing set with novel classes unseen during training. Each task contains some unlabeled samples (query shots) and one labeled sample (support shot). We evaluate our TP-VAE model performance over the few-shot tasks.

Meta learning based method is a representative paradigm to Few-Shot Learning (FSL) solutions. In the meta learning settings, we view the meta training set as a series of independent tasks, and we use support shots and query shots (from the testing set) to simulate generalization difficulties during test times. Many existing approaches  \cite{lee2019meta,tian2020rethinking,ziko2020laplacian,boudiaf2020transductive} view the meta training (base-training) stage as a feature extraction (embedding) process, which projects the data from the raw data space to the latent feature space. In the feature space, we can easily classify the query shots. In the stage of meta training, a backbone neural network will be trained as a feature extractor.
% During meta testing, \cite{lee2019meta} use a support vector machine and\cite{tian2020rethinking} use a linear classifier to classify the images.

Several recent works devote to transductive inference for FSL,e.g., \cite{ziko2020laplacian,boudiaf2020transductive,hu2020empirical,Dhillon2020A,CAN,liu2019fewTPN}, among others. In the transductive settings, the unlabeled query shots participate in training during the meta testing phase instead of only one or a few labeled support shots used like inductive methods.

In the transductive few-shot learning scenarios, query shots supply sufficient information to make the classification results better. However, the existing methods of transductive FSL only consider one situation: the number of query shots for each class is the same, e.g., 15 query shots per class. This setting contains a strong prior condition that the proportion of query shots in each class is the same, but this prior is limited for many situations in the real world. For example, in the field of medical imaging, the number of medical images for each type of disease is different. Through experiments, we discover that some existing methods will suffer performance when the proportion of query shots in each class is different. To cope with this problem, we propose a transductive few-shot method combined with a Task-Prior Conditional Variational Auto-Encoder, named TP-VAE. In our method, the task-level prior provides a regularization, constraining the proportion of query shots in each class. The objective 
consists of two parts: a standard cross-entropy for the support shots, and a task-prior conditional evidence lower bound (ELBO). 

% the contribution of this paper include three part as follow:
% \begin{itemize}
%             \item We propose TP-VAE for transductive few-shot learning. Our method optimizes the task-prior level conditional ELBO of the query shots and the standard cross-entropy for support shots.
%           \item our method can maintain good performance regardless of whether the proportion of query shots in each category is the same. Interestingly, when the query shots are not uniform, we always can obtain better performance in the case of one-shot scenario.
%             \item We conduct extensive experiments and show that in the standard transductive few-shot scenario, with uniformed query shots for each task like other methods do, TP-VAE outperforms state-of-the-art methods in various datasets and networks. During the meta training stage, we just use simple cross-entropy training on the base classes, without complex meta learning schemes. Compared with the best-performing methods, our method obtains between $2 \%$ and $3 \%$ of improvement in accuracy in the one-shot scenario, meanwhile, in the five-shot scenario we also obtain a small improvement.
%         \end{itemize}
Our contributions are summarized as below:
\begin{enumerate}
    \item We propose TP-VAE for transductive few-shot learning. Our method optimizes the Task-Prior Conditional ELBO of the query shots and the standard cross-entropy of the support shots. We confirm through ablation experiments that each module in our model contributes to the classification results.
           \item We propose a more challenging scenario for few-shot learning (i.e. nonuniform few-shot learning). For fair comparison, we simulate two different nonuniform few-shot scenarios. The experiments show that our method can maintain high performance in this more challenging situation.
            \item We conduct extensive experiments in standard few-shot scenario, and the results show that in the standard transductive few-shot scenario, TP-VAE outperforms the state-of-the-art methods in various datasets and networks. To our best of knowledge, the TP-VAE has increased an accuracy of 1-shot by about 3\%. Meanwhile, we also obtain a slight improvement in 5-shot scenario. 
\end{enumerate}
    \section{Related Work}
    
    \subsection{Meta Learning}
      A large body of previous work addresses the few-shots problem in the meta learning framework, which divides the training process into two part, i.e. meta training and meta testing. We learn the model parameters in the meta training stage and enable the model to quickly generalize to new tasks by sharing learned knowledge with new tasks. Existing methods of meta learning can be generally divided into three categories. Our method is related to two of them: (1) \cite{finn2017model,rusu2018meta,rajeswaran2019meta,zintgraf2019fast,jamal2019task} propose a series of optimization-based methods.  MAML \cite{finn2017model} learns a general initialization model by optimizing the second-order gradient so that when the model faces new tasks, it can converge after a few iterations. \cite{yoon2018bayesian} explains MAML in the perspective of Bayesian inference. Reptile \cite{nichol2018reptile} points out that MAML consumes a lot of resources to calculate the second-order Hessian matrix. They propose a fast algorithm that calculates the first-order Jacobi matrix rather than the Hessian matrix to overcome this difficulty in computational cost.
    (2) Metric based methods \cite{vinyals2016matching,snell2017prototypical,sung2018learning,yoon2019tapnet,xu2020attentional,CAN,allen2019infinite} map the raw images into a latent feature space, and classify query shots by measuring the distance from support shots. Matching network \cite{vinyals2016matching} learns to map small labeled support shots and an unlabeled query shot to their own labels. Prototypical network \cite{snell2017prototypical} maps the sample data in each class into a feature space and extract their "mean" to represent the prototype of each class. \cite{allen2019infinite} proposes an infinite mixture prototype to generalized complex data distributions.
\subsection{Transductive Few-Shot Learning}
Many previous works \cite{ziko2020laplacian,boudiaf2020transductive,hu2020empirical,Dhillon2020A,CAN,liu2019fewTPN,hu2021leveraging,qiao2019transductive,wang2020instance,guo2020attentive,yang2020dpgn} show that transductive few-shot methods can always  outperform inductive few-shot methods due to that transductive methods making full use of the information provided by the query shots. TPN \cite{liu2018learning} proposes a graphical model, which learns to propagate labels between data instances for unseen classes via episodic meta learning. CSPN \cite{liu2020prototype} analyzes the inner bias between query shot and support set and decreased this bias to improve classification accuracy. LaplacianShot few-shot \cite{ziko2020laplacian} uses spectral clustering and propose the Laplacian regularization method. TIM \cite{boudiaf2020transductive} maximizes the mutual information between query feature and their predict label, and proposes a very fast optimize method for their TIM named TIM-ADM. 

\subsection{Deep Conditional Variational AutoEncoder}
Variational Auto-Encoder (VAE) \cite{kingma2013auto} is a famous paradigm of deep generative models. On the basis of VAE, many different forms of conditional VAE \cite{sohn2015learning,yu2020draft,manduchi2021deep} are applied to a large number of applications in different tasks such as image generation, text generation, and image classification.  \cite{sohn2015learning} conditions on the label message, and \cite{manduchi2021deep} conditions on prior clustering preferences with pairwise constrain. Inspired by this, we propose a VAE conditioned on support set in the transductive few-shot scenario. The support set contains the mean massage of distribution for each cluster (the centroid of each cluster).

\section{Task-Prior Variational Auto-encoder}
\subsection{ Problem Formulation}

 Following \cite{ziko2020laplacian,boudiaf2020transductive} , we are given a labeled training set, $\mathcal D_\textrm{base}=\{x_i,c_i\}_{i=1}^{N_\textrm{base}}$, where $x_i$ denotes raw data and $c_i$ denotes the label of $x_i$ with one-hot encode. This training set is consistent with meta training set. $\mathcal C_\textrm{base}$ is the set of labels for $\mathcal D_\textrm{base}$. To generate few-shot tasks, we have another dataset  $\mathcal D_\textrm{test}=\{x_i,c_i\}_{i=1}^{N_{test}}$, and $\mathcal C_\textrm{test}$ is the set of labels for $\mathcal D_\textrm{test}$. It is worth noting that $\mathcal C_\textrm{test}\cap\mathcal C_\textrm{base}=\emptyset$. For each N-way K-shot task, we construct a labeled support set ${\mathcal S}=\cup_{n=1}^N\{x_i^n,c_i^n\}_{i=1}^{k}$ and an unlabeled query set ${\mathcal Q}=\cup_{n=1}^N\{x_i^n\}_{i=1}^{N_q}$. Both support set ${\mathcal S}$ and query set ${\mathcal Q}$ are the subset of $\mathcal D_\textrm{test}$. The number of query shots for each class will be adjusted according to different tasks, which will be discussed in the experimental part. \\
In the few-shot scenario, $g_\theta(\cdot)$ is the decoder network, and $f_{\phi}(\cdot)$ denotes the fixed feature-extractor already trained on $\mathcal D_\textrm{base}$ with standard cross-entropy. This training is without any complex meta training or specific sampling schemes. $\mathcal X$ denote the random variables associated with the raw image data, and $ z_i=f_\phi(x_i)\in\mathbb R^d$ denote a continuous latent embedding features.
$\mathcal C\in\{1,\dots, K\}$ correspond to the real labels of the data, and $\mathcal Y\in\{1,\dots, K\}$ correspond to the inference predict labels.

We propose our objective function as follows:
\begin{equation}\label{loss function}
\min\limits_{\phi,\psi}\  \mathcal{L}_{\phi, \psi}^\textrm{ce}+ \mathcal{L}_{\phi, \psi}^\textrm{tpce}
\end{equation}
The first part of Eq.\ref{loss function} is a standard cross-entropy of support features :
\begin{equation}
\label{cross-entropy}
   \mathcal{L}^\textrm{ce}=\frac 1 L\sum_{l=1}^L\sum_{s=1}^S\sum_{k=1}^Kc_{sk}\log p_\psi(y_{s}=k|z_{s}^l)
\end{equation}

The second part of Eq.\ref{loss function} is a Task-Prior Conditional ELBO:
\begin{equation}
\label{pce_big}
\begin{aligned}
\mathcal{L}^\textrm{tpce}&=\frac 1 L\bigg{[}\underbrace{\sum_{k=1}^K[\sum_{i=1}^Np_\psi(y_i=k|z_i^l,S)]\cdot\log\left(\sum_{i=1}^N \frac{p(y_i)}{p_\psi(y_i=k|z_i^l,S)}\right)}_{\mathcal{L}_{tp}: \text{task-prior KL divergence}}\\
&+\sum_{i=1}^N\sum_{k=1}^K p_\psi(y_i=k|z_i^l,S)\log p(z_i^l|{y_i=k})
+\sum_{l=1}^L\sum_{i=1}^N \log p_\theta(q_i|z_i^l)
\bigg{]}
\end{aligned}
\end{equation}
In the objective function, $z_i^l$  are the embedding features samples.

The distribution of each random variable can be summarized as follows:
\begin{align}
p_\theta(q_i|z_i)&\sim \mathcal N(\mu_{q_i},\sigma_{q_i}\mathbb I)
\\
\label{post_z}p(z_i|y_i=k)&\sim \mathcal N(\mu_{y_i=k},\sigma_{y_i=k}^2\mathbb I)\\
\label{post_y}p_\psi(y_i=k|z_i)&\sim Cat(k)
\end{align}

Where $[\mu_{q_i},\sigma_{q_i}^2\mathbb I]$ is obtained by decoder $g_\theta(\cdot)$, and  $[\mu_{y_i=k},\sigma_{y_i=k}^2\mathbb I]$ represents the mean and variance of the Gaussian distribution corresponding to $k\text{th}$ cluster in feature space when $y_i=k$. $\psi$ represent a classifier initialed by support embedding features $z_s$. We do inference about the posterior distribution $p_\psi(y_i|z_i)$ by prototype nearest-neighbor classification \cite{wang2019simpleshot}. The specific inference process is as follows:

Initialize the classifier:
\begin{equation}
\label{init_clas}
\psi_k^{(0)}=\frac{\sum_{i\in s}c_{ik}z_i}{\sum_{i\in s}c_{ik}}
\end{equation}

Then the mean of $k\text{th}$ Gaussian distribution equals $\psi_k$ , i.e. $\psi_k=\mu_{y_i=k}$.

After prototype nearest-neighbor classification \cite{wang2019simpleshot}, we can obtain the following relation: 
\begin{align}
p(z_i|y_i=k)&\propto \exp({\frac{-(z_i-\psi_k)^2}{\sigma^2}})\\
\label{post1}
p_\psi(y_i=k|z_i) &\propto \frac{\exp({\frac{-(z_i-\psi_k)^2}{\sigma^2}})}{\sum\limits_{j=1}^k \exp({\frac{-(z_i-\psi_j)^2}{\sigma^2})}}
\end{align}

\begin{figure}[!tb]
            \centering
            \includegraphics[scale=0.28]{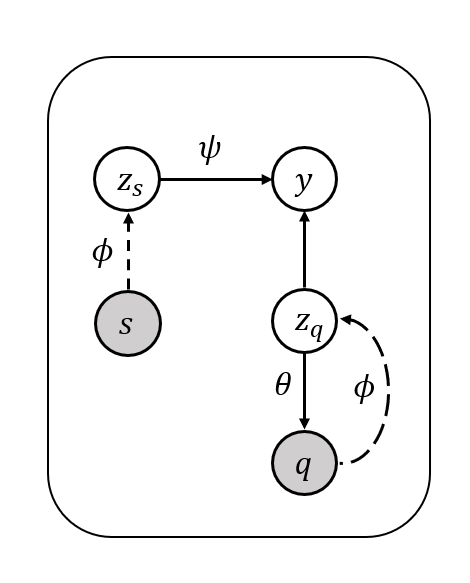}
            \caption{Graphical model for TP-VAE. The dashed line denotes the variational approximation process. The solid lines denote generate process and prototype nearest-neighbor inference process }
            \label{fig:graph model}
        \end{figure}
        
About the prior $p(y_i)$, We take the result of the first classification result by untrained classifier $\psi_k$ (support feature) as our prior:
\begin{align}
\label{prior}
p(y_i)=p_\psi^{(1)}(y_i=k|z_i)
\end{align}

The graphical illustration of TP-VAE is depicted in Fig. \ref{fig:graph model}, and we detail algorithm of TP-VAE in Algorithm \ref{algorithm}.
\subsection{Task-Prior Conditional ELBO  } 
\begin{algorithm}[!tb]
            \caption{Proposed Algorithm for TP-VAE}
            \label{algorithm}
            \KwData{$\mathcal S$, $\mathcal Q$}
            \KwOut{posterior distribution $p_\psi(y_i=k|z_i)$  }
            initialize $\theta$ randomly\;
            initialize $\psi$ using (\ref{init_clas})\;
            computer $\phi$ at the stage of base-training\;
            compute$\ p(y_i) \text{using}\ (\ref{prior})$\;
            \While{not done }{
                sampling $z_i^l\ \text{from} \ q_{\phi}(z)$\;
                compute $p_\psi(y_i=k|z_i)\  \text{using}\ (\ref{post1})$\;
                 compute $\mathcal{L}_{tpce}\  \text{using}\ (\ref{pce_big})$\;
                optimize $\mathcal{L}_{tpce}\  \text{using}\ \text{SGD}$\;
        }
      
    \end{algorithm}

The log-likelihood of the query data can be written as follows:
\begin{align}
% \begin{aligned}
\log p(\mathcal Q|\mathcal S) &=\log \frac{p(\mathcal Q,Z,Y|\mathcal S)}{p(Z,Y|\mathcal Q,\mathcal S)} \\
&=\mathbb E_{q_{\phi,\psi}(Z,Y|\mathcal Q,\mathcal S)}\log \frac{p(\mathcal Q,Z,Y|\mathcal S)}{p(Z,Y|\mathcal Q,\mathcal S)}\\\label{l_pc}
&=\mathbb E_{q_{\phi,\psi}(Z,Y|\mathcal Q,\mathcal S)} \log\frac{p(\mathcal Q,Z,Y|\mathcal S)}{q_{\phi,\psi}(Z,Y|\mathcal Q,\mathcal S)}\\&+\mathbb E_{q_{\phi,\psi}(Z,Y|\mathcal Q,\mathcal S)}\log\frac{q_{\phi,\psi}(Z,Y|\mathcal Q,\mathcal S)}{p(Z,Y|\mathcal Q,\mathcal S)}\notag
% \end{aligned}
\end{align}

Where $Z=\{z_i\}_{i=1}^N$ are the latent embedding features associated with the query shots. $Y=\{y_i\}_{i=1}^N\in\{1,2\cdots, k\}$ are the latent predict label. The first term of Eq.\ref{l_pc} is a conditional ELBO of the marginal likelihood. The second term is a KL divergence. When $q_{\phi,\psi}(Z,Y|\mathcal Q,\mathcal S)=p(Z,Y|\mathcal Q,\mathcal S)$ , the ELBO equals the marginal likelihood. 

The conditional ELBO can be written as follows: 
\begin{align}
\mathcal{L}^{cpe}&=
\mathbb E_{q_{\phi,\psi}(Z,Y|\mathcal Q,\mathcal S)} \log\frac{p(\mathcal Q,Z,Y|\mathcal S)}{q_{\phi,\psi}(Z,Y|\mathcal Q,\mathcal S)}\\
&=\mathbb E_{q_{\phi,\psi}(Z,Y|\mathcal Q,\mathcal S)} \log p_\theta(\mathcal Q|Z)+\mathbb E_{q_{\phi,\psi}(Z,Y|\mathcal Q,\mathcal S)}\log \frac{p(Z,Y|S)}{q_{\phi,\psi}(Z,Y|\mathcal Q,\mathcal S)}\\
\label{L_cpe}&=\mathbb E_{q_\phi(Z|\mathcal Q)}\log p_\theta(\mathcal Q|Z) +\mathbb E_{q_\phi(Z|\mathcal Q)p_\psi(Y|Z,\mathcal S)}\log p(Z|Y)\\&+\mathbb E_{q_\phi(Z|\mathcal Q)p_\psi(Y|Z,\mathcal S)}\log \frac {p(Y)}{p_\psi(Y|Z,\mathcal S)}
-\mathbb E_{q_\phi(Z|\mathcal Q)}\log q_\phi(Z|\mathcal Q)\notag
\end{align}

When  $q_\phi(Z|\mathcal Q)$ is Gaussian, the last part of Eq.\ref{L_cpe} can be factorized as follows:
\begin{align}
\label{constant}
\mathbb E_{q_\phi(Z|\mathcal Q)}\log q_\phi(Z|\mathcal Q)
  =-\frac{D}{2}\log(2\pi)-\frac{1}{2}\sum_{d=1}^D(1+\log \sigma_d^2)
  \end{align}

Where $d$ represents the dimension of the Gaussian distribution. When we set $\sigma_j^2$ to a constant, Eq.\ref{constant} becomes a constant. Then we remove it from the objective function.

Given $q_\phi(Z|\mathcal Q)p_\psi(Y|Z,\mathcal S)=\prod \limits_{i=1}^N q_\phi(z_i|q_i)p_\psi(y_i|z_i,s_i)$, we use the Monte Carlo sampling method to sample $z_i$ from $q_\phi(z_i|q_i)      $ , $\mathcal{L}^{cpe}$ can be factorized as follows:
\begin{equation}\begin{aligned}
\label{chaifen}
\mathcal{L}^{cpe}&=\frac 1 L\sum_{l=1}^L[\sum_{i=1}^N \log p_\theta(q_i|z_i^l)+\sum_{i=1}^N\sum_{k=1}^K p_\psi(y_i=k|z_i^l,S)\log p(z_i^l|{y_i=k})\\
&+\sum_{i=1}^N\sum_{k=1}^Kp_\psi(y_i=k|z_i^l,S)\log\frac{p(y_i)}{p_\psi(y_i=k|z_i^l,S)}]
\end{aligned}
\end{equation}

The third part of the Eq.\ref{chaifen} is a KL divergence between the posterior distribution  $p_\psi(y_i|z_i)$ and prior $p(y_i)$. As mentioned above, we calculate prior from Eq.\ref{prior}. However, if we just optimize the KL divergence at the sample level (every query shot is regularized by the prior), the classification result will stay in the first classification result and can't converge to a better result.
For our model converge to a better results, we use the Jensen inequality to lighten the constraint of the sample level KL divergence. Then we obtain a task level KL divergence: proportion of query shots in each class are regularized by the prior (e.g., if the category of cats in the first classification accounts for 30\% of the total categories, We should regularize the proportion of cat category to 30\% in the prior). Then we can obtain the Task-Prior Conditional ELBO as follows: 
\begin{equation}\begin{aligned}
\label{lpce}
\mathcal{L}^{cpe}&\ge\frac 1 L\bigg{[}\sum_{l=1}^L\sum_{i=1}^N \log p_\theta(q_i|z_i^l)+\sum_{i=1}^N\sum_{k=1}^K p_\psi(y_i=k|z_i^l,S)\log p(z_i^l|{y_i=k})\\
&+\sum_{k=1}^K[\sum_{i=1}^Np_\psi(y_i=k|z_i^l,S)]\cdot\log(\sum_{i=1}^N \frac{p(y_i)}{p_\psi(y_i=k|z_i^l,S)})\bigg{]}=\mathcal{L}^\textrm{tpce}
\end{aligned}
\end{equation}

\section{Experiments}
This section provides our experiment setup details for the few-shots image classification. We report the average of accuracies evaluated over 1000 episodes in all our experiments.

\subsection{Datasets}

 We use three standard benchmarks for few-shot image classification: \textit{mini}-ImageNet \cite{vinyals2016matching}, \textit{tiered}-ImageNet \cite{ren2018meta}, and cub-200-2011 \cite{wah2011caltech}.
 
 Mini-ImageNet has 60000 color images with 100 classes. Following previous work \cite{ravi2016optimization,wang2019simpleshot}, we divide the dataset into 64 bases, 16 validations and 20 test classes.
 
 Tiered-ImageNet has large data with 608 classes.
 Following \cite{ren2018meta}, we divide the dataset into 351 bases, 97 validations and 160 test classes.
 
 CUB-200-2011 is a fine-grained dataset. We follow \cite{chen2019closer} splitting the dataset into 100 bases, 50 validations and 50 test classes for the experiments.
\subsection{Hyper-Parameters}
We keep the hyper-parameter fixed across all the models and datasets to make our experiments easier to implement. The label-smoothing parameter is set to 0.1 during the meta training stage. When we implement our TP-VAE, the learning rate of the SGD optimizer is set to $1\times10^{-1}$ with momentum 0. The number of samples $l$ is set to 1. Inverse of variance always be seen as a temperature parameter $\tau$ in many contrastive learning and few-shot learning scenarios \cite{boudiaf2020transductive,wang2019simpleshot,li2020prototypical}. We choose appropriate temperature parameters in the ablation experiments.
\subsection{Base-Training Procedure}
We evaluate TP-VAE on two different backbone networks models as the feature extractor $f_\phi$:
ResNet-18 \cite{he2016deep} is the most common neural networks used in deep learning, which has 8 basic residual blocks. WRN-28 \cite{zagoruyko2016wide} has more convolutional layers and feature planes, which often achieve better results than ResNet-18 as a feature extractor. As for the decoder neural network $g_\theta$, we use a simple Conv-4 network \cite{vinyals2016matching}. 

For a fair comparison with previous works, our feature extractor is trained following LaplacianShot and TIM \cite{ziko2020laplacian,boudiaf2020transductive}. We train the feature extractor using the standard cross-entropy with label smoothing. It is worth mentioning that the base training stage doesn't involve any other complex meta learning or episodic-training strategy. We use the SGD optimizer to train the feature extractor with an early stopping strategy, and we set 256 batch sizes for ResNet-18 and 128 batch sizes for WRN-28. About data augment, we use random cropping, color jitter and random horizontal flipping like \cite{ziko2020laplacian,boudiaf2020transductive}.

\subsection{TP-VAE Implement}
\begin{figure}[t]
            \centering
            \includegraphics[scale=0.24]{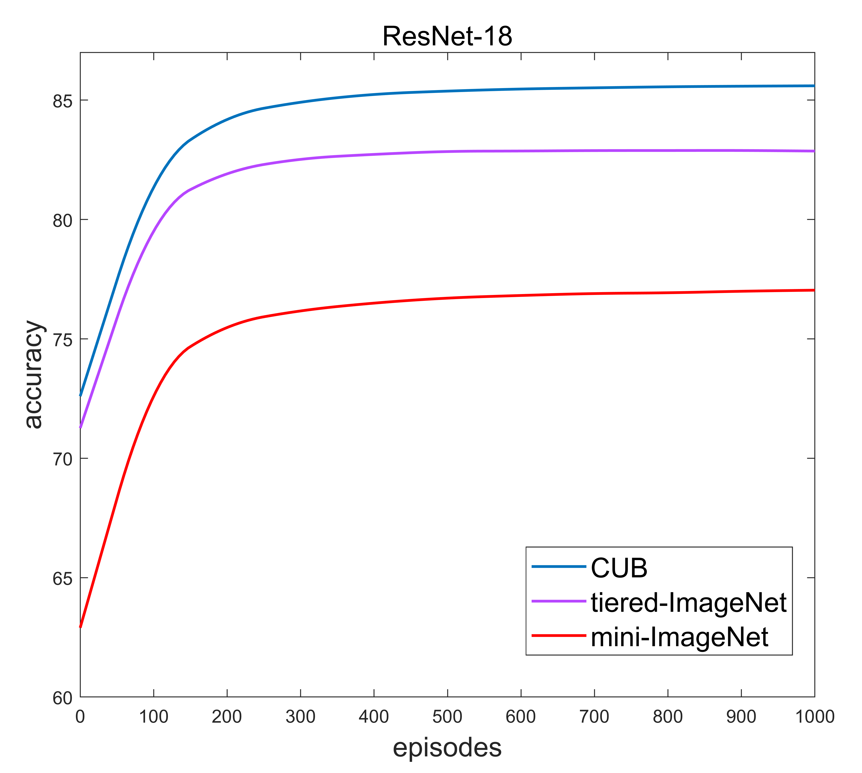}
            \caption{The accuracy improvement trend during the training process
            % on \textit{mini}-ImageNet, \textit{tiered}-ImageNet and CUB-200-2011 with ResNet-18}
            }
            \label{fig:convergence}
        \end{figure}

We implement our TP-VAE  as described in Algorithm \ref{algorithm}. We record the training process of TP-VAE on three benchmark datasets with ResNet-18. The results are recorded in Fig. \ref{fig:convergence}. It can be observed that our method can achieve high performance at 200 episodes and converge within 1000 episodes on all datasets.

\subsection{Comparison Results with the State-of-The-Art Methods}
\begin{table}[!tb]
    \centering
    \small
     \caption{Comparison to the state-of-the-art methods on \textit{mini}-ImageNet and \textit{tiered}-Imagenet. The best performances are shown in \textbf{bold}, and "-" signifies the result is unavailable}
    \begin{tabular}{lcccccc}
    \hline\noalign{\smallskip}
         & & &\multicolumn{2}{c}{\textbf{\textit{mini}-ImageNet}} & \multicolumn{2}{c}{\textbf{\textit{tiered}-ImageNet}}  \\
            \cline{4-7}\noalign{\smallskip}
             Method & Transductive & Backbone & 1-shot & 5-shot & 1-shot & 5-shot\\
             
             \hline\noalign{\smallskip}
              MAML \cite{finn2017model}& {\xmark} & ResNet-18 & 49.6 & 65.7 & - & - \\
              \noalign{\smallskip}
              MatchNet \cite{vinyals2016matching}&  {\xmark}& ResNet-18 & 52.9 & 68.9 &  - & - \\ 
              \noalign{\smallskip}
              ProtoNet\cite{snell2017prototypical}&{\xmark} & ResNet-18 & 54.2 & 73.4 &  - & -  \\
              \noalign{\smallskip}
              TPN \cite{liu2018learning} &{\cmark}& ResNet-12 & 59.5 & 75.7 & - & - \\
              \noalign{\smallskip}
               TEAM \cite{qiao2019transductive}&{\cmark} & ResNet-18 & 60.1 & 75.9 & - & - \\
               \noalign{\smallskip}
                 MTL \cite{sun2019meta} &{\cmark} & ResNet-12 & 61.2 & 75.5 & - & -\\
                 \noalign{\smallskip}
                  Neg-cosine \cite{liu2020negative}&{\xmark} & ResNet-18 & 62.3 & 80.9 & - & -  \\
                  \noalign{\smallskip}
              MetaOpt \cite{lee2019meta} &{\xmark}& ResNet-12 & 62.6 & 78.6 & 66.0 & 81.6 \\
              \noalign{\smallskip}
                SimpleShot \cite{wang2019simpleshot} &{\xmark} & ResNet-18 & 62.9 & 80.0 & 68.9 & 84.6  \\
                \noalign{\smallskip}
             Distill \cite{tian2020rethinking} &{\xmark} & ResNet-12 & 64.8 & 82.1 & 71.5 & 86.0\\
             \noalign{\smallskip}
              CAN+T \cite{CAN}& {\cmark}& ResNet-12 & 67.2 & 80.6 & 73.2 & 84.9 \\
              \noalign{\smallskip}
               LaplacianShot \cite{ziko2020laplacian}&{\cmark} & ResNet-18 & 72.1 & 82.3 & 79.0 & 86.4 \\
               \noalign{\smallskip}
               TIM\cite{boudiaf2020transductive} & {\cmark} & ResNet-18 & 73.9 & 85.0 & 79.9 & 88.5\\
               \noalign{\smallskip}
               TP-VAE (ours) &{\cmark} & ResNet-18 & \textbf{76.9} & \textbf{85.2} &\textbf{ 82.9} & \textbf{89.1}  \\
               
             \hline\noalign{\smallskip}
              LEO \cite{rusu2018meta}& {\xmark} & WRN28-10 & 61.8 & 77.6 & 66.3 & 81.4 \\
              \noalign{\smallskip}
               AWGIM \cite{guo2020attentive}&{\cmark}  & WRN28-10 & 63.1 & 78.4 & 67.7 & 82.8\\
               \noalign{\smallskip}
               SimpleShot \cite{wang2019simpleshot}&{\xmark} & WRN28-10 & 63.5 & 80.3 & 69.8 & 85.3\\
               \noalign{\smallskip}
                MatchNet \cite{vinyals2016matching}&{\xmark} & WRN28-10 & 64.0 & 76.3 & - & -\\
                \noalign{\smallskip}
                FEAT \cite{ye2019learning} &{\xmark} & WRN28-10 & 65.1 & 81.1 & 70.4 & 84.4 \\
                \noalign{\smallskip}
               Ent-min \cite{Dhillon2020A} & {\cmark} & WRN28-10 & 65.7 & 78.4 & 73.3 & 85.5 \\
               \noalign{\smallskip}
                SIB  \cite{hu2020empirical} &{\cmark}  & WRN28-10 & 70.0 & 79.2 & - & - \\
                \noalign{\smallskip}
                 LaplacianShot  \cite{ziko2020laplacian} &{\cmark}  & WRN28-10 & 74.9 & 84.1 & 80.2 & 87.6 \\
                 \noalign{\smallskip}
                 TIM\cite{boudiaf2020transductive} & {\cmark} & WRN28-10 & 77.8 & \textbf{87.4} & 82.1 & 89.8\\
                 \noalign{\smallskip}
                  TP-VAE (ours) &{\cmark} & WRN28-10 & \textbf{79.8} & 87.2 &\textbf{ 84.3} & \textbf{90.2}  \\
                 \hline\noalign{\smallskip}
    \end{tabular}
    \label{tab:sota}
\end{table}

\begin{table}[!tb]
    \centering
    \small
     \caption{Comparison to the state-of-the-art methods on CUB. The best performances are shown in \textbf{bold}, and "-" signifies the result is unavailable}
\begin{tabular}{lcccc}
            \hline\noalign{\smallskip}
         & & &\multicolumn{2}{c}{\textbf{CUB-200-2011}}  \\
            \cline{4-5}\noalign{\smallskip}
             Method & Transductive & Backbone & 1-shot & 5-shot \\
             
             \hline\noalign{\smallskip}
              MAML \cite{finn2017model}& {\xmark} & ResNet-18 & 68.4 & 83.5  \\
              \noalign{\smallskip}
              MatchNet \cite{vinyals2016matching}&  {\xmark}& ResNet-18 & 73.5 & 84.5\\ 
              \noalign{\smallskip}
              ProtoNet\cite{snell2017prototypical}&{\xmark} & ResNet-18 & 73.0 & 86.6  \\
              \noalign{\smallskip}
              TPN \cite{liu2018learning} &{\cmark}& ResNet-12 & - & -  \\
              \noalign{\smallskip}
               TEAM \cite{qiao2019transductive}&{\cmark} & ResNet-18 & - & - \\
               \noalign{\smallskip}
                 MTL \cite{sun2019meta} &{\cmark} & ResNet-12 & - & - \\
                 \noalign{\smallskip}
                  Neg-cosine \cite{liu2020negative}&{\xmark} & ResNet-18 & 72.7 & 89.4  \\
                  \noalign{\smallskip}
              MetaOpt \cite{lee2019meta} &{\xmark}& ResNet-12 & - & -  \\
              \noalign{\smallskip}
                SimpleShot \cite{wang2019simpleshot} &{\xmark} & ResNet-18 & 68.9 & 84.0  \\
                \noalign{\smallskip}
             Distill \cite{tian2020rethinking} &{\xmark} & ResNet-12 & - & - \\
             \noalign{\smallskip}
              CAN+T \cite{CAN}& {\cmark}& ResNet-12 & - & -  \\
              \noalign{\smallskip}
               LaplacianShot \cite{ziko2020laplacian}&{\cmark} & ResNet-18 & 81.0 & 88.7\\
               \noalign{\smallskip}
               TIM\cite{boudiaf2020transductive} & {\cmark} & ResNet-18 & 82.2 & 90.8\\
               \noalign{\smallskip}
               TP-VAE (ours) &{\cmark} & ResNet-18 & \textbf{85.6} & \textbf{91.0}   \\
               \hline\noalign{\smallskip}
 \end{tabular}
    \label{tab:sota_cub}
\end{table}

We evaluate our TP-VAE in the standard 5-way 1-shot and 5-way 5-shot scenarios with {\it{15 query shot for each class }} (total 75 query shots) on the Mini-ImageNet, tired-ImageNet, and CUB-2011. The results are reported in Table \ref{tab:sota} and Table \ref{tab:sota_cub}. TP-VAE can achieve the state-of-the-art performances by a large margin, with different network models and datasets. For instance, compared 
with LaplacianShot and TIM, TP-VAE outperforms 5\% and 3\% respectively in 1-shot scenario.
To illustrate our model can be widely applied to few-shot learning, not just limited to the 5-way scenario, we also do experiments on more challenging 10-way and 20-way scenarios. The results are reported in Table \ref{10way}. Our method also achieves the state-of-the-art in this more challenging scenario.

\begin{table}[!tb]
 \caption{Test results for 10-way and 20-way scenarios on \textit{mini}-ImageNet with ResNet-18 }
    \centering
    \small
     \begin{tabular}{lcccc}
     \hline\noalign{\smallskip}
     &\multicolumn{2}{c}{\textbf{10way}} & \multicolumn{2}{c}{\textbf{20way}} \\
     \cline{1-5}\noalign{\smallskip}
     Method&1-shot&5-shot&1-shot&5-shot\\
     \noalign{\smallskip}
     MatchNet\cite{vinyals2016matching}&-&52.3&-&36.8\\
     \noalign{\smallskip}
     Baseline\cite{chen2019closer}&-&55.0&-&42.0\\
     \noalign{\smallskip}
      ProtoNet\cite{snell2017prototypical}&-&59.2&-&45\\
      \noalign{\smallskip}
     SimpleShot\cite{wang2019simpleshot}&45.1&68.1&32.4&55.4\\
     \noalign{\smallskip}
     TIM\cite{boudiaf2020transductive}&56.1&72.8&39.3&\textbf{59.5}\\
     \noalign{\smallskip}
      TP-VAE (ours)&\textbf{57.7}&\textbf{73.2}&\textbf{39.5}&59.4\\
    \hline
         \end{tabular}
    \label{10way}
\end{table}

\subsection{TP-VAE for Nonuniform Few-shot Image Classification }

% Except for the number of query shots in each class, the settings of this part are the same as Uniform Few-shot.
In order to simulate the situation that the number of samples of different categories in the real world is not uniform, we simulate two different situations: 

1. The proportion of query shots in each class is slightly different (we set the number of query shots to [20, 20, 10, 10, 15]). 

2. The number of query shots in one class is extremely small ([19, 19, 18, 18 ,1]).

We compare our method with TIM \cite{boudiaf2020transductive} and LaplacianShot \cite{ziko2020laplacian} on \textit{mini}-ImageNet, \textit{tiered}-ImageNet, and CUB-200-2011 with ResNet-18. The main results are shown in Table \ref{tab:ununiform} and Table \ref{tab:uniform}. In these two simulation scenarios, the performance of TIM is greatly affected, which indicates that TIM method potentially exploits the strong uniform prior condition. In comparison, our method can maintain high performance in all scenarios. Specifically, compared with TIM and LaplacianShot fairly, TP-VAE brings about 5\% improvement in the nonuniform 1-shot scenario and about 4\% improvement in the nonuniform 5-shot scenario. 
% We again demonstrate the superiority of our method in transductive few-shot learning.  
\begin{table}[!tb]
 \caption{Test results for extremely non-uniformed few-shot scenario with ResNet-18  }
    \centering
    \small
     \begin{tabular}{lcccccc}
     \hline\noalign{\smallskip}
     &\multicolumn{2}{c}{\textbf{\textit{mini}-ImageNet}} & \multicolumn{2}{c}{\textbf{\textit{tiered}-ImageNet}} & \multicolumn{2}{c}{\textbf{CUB-200-2011}}\\
     \cline{1-7}\noalign{\smallskip}
     Method&1-shot&5-shot&1-shot&5-shot&1-shot&5-shot\\
     \noalign{\smallskip}
     LaplacianShot&69.3&80.2&76.1&84.7&78.0&87.0\\
     \noalign{\smallskip}
     TIM&64.0&72.6&68.6&75.2&70.5&77.2\\
     \noalign{\smallskip}
     TP-VAE&\textbf{74.1}&\textbf{85.5}&\textbf{81.7}&\textbf{88.9}&\textbf{84.5}&\textbf{91.0}\\
     \hline
         \end{tabular}
    \label{tab:ununiform}
\end{table}

\begin{table}[!tb]
 \caption{Test results for slightly non-uniformed few-shot scenario with ResNet-18 }
    \centering
    \small
     \begin{tabular}{lcccccc}
     \hline\noalign{\smallskip}
     &\multicolumn{2}{c}{\textbf{\textit{mini}-ImageNet}} & \multicolumn{2}{c}{\textbf{\textit{tiered}-ImageNet}} & \multicolumn{2}{c}{\textbf{CUB-200-2011}}\\
     \cline{1-7}\noalign{\smallskip}
     Method&1-shot&5-shot&1-shot&5-shot&1-shot&5-shot\\
     \noalign{\smallskip}
     LaplacianShot&69.3&81.4&76.6&85.8&79.0&87.9\\
     \noalign{\smallskip}
     TIM&62.7&71.2&67.0&73.6&68.7&75.6\\
     \noalign{\smallskip}
     TP-VAE&\textbf{75.1}&\textbf{85.2}&\textbf{81.6}&\textbf{88.9}&\textbf{84.5}&\textbf{91.0}\\
 \hline\noalign{\smallskip}
         \end{tabular}
    \label{tab:uniform}
\end{table}
% \subsection{Convergence Analysis}
% To verify our method can quickly converge, we record the training process on three benchmark datasets with ResNet-18 during meta-testing stage. The results are shown in figure \ref{fig:convergence}. Our method can achieve high performance at 200 episodes and converge within 1000 episodes on all datasets.
%  \begin{figure}[H]
%             \centering
%             \includegraphics[scale=0.25]{fig10.png}
%             \caption{Training process of TP-VAE on mini-ImageNet, tiered-ImageNet and CUB-200-2011 with ResNet-18  }
%             \label{fig:convergence}
%         \end{figure}

\subsection{Ablation Study}
\textbf{Effects of the task-prior regularization}:
Our objective function consists of two parts: $\mathcal{L}^{\textrm{ce}}\ $and $\mathcal{L}^{\textrm{tpce}}\ $. To evaluate the effects of the task-prior regularization, we divide $\mathcal{L}^{\textrm{tpce}}\ $ into two parts, the task-prior KL divergence is recorded as $\mathcal{L}^{\textrm{tp}}$, and the remainder is denoted as $\mathcal{L}^{\textrm{re}}\ $. The result reported in Table \ref{tab:ablation}. All the components of our objective function contribute to the final result: the cross-entropy of the support shot allows us make full use of the information contained in support shots and its labels. The remainder part $\mathcal{L}^\textrm{re}$ makes the query shots into a class more confident. The task-prior KL divergence regular constraints the objective function so that the final result is not all query shots are classified into one class.
\begin{table}[!b]
 \caption{Ablation study on the effect of each component in our objective function with ResNet-18}
    \centering
    \small
     \begin{tabular}{lcccccc}
     \hline\noalign{\smallskip}
     &\multicolumn{2}{c}{\textbf{\textit{mini}-ImageNet}} & \multicolumn{2}{c}{\textbf{\textit{tiered}-ImageNet}} & \multicolumn{2}{c}{\textbf{CUB-200-2011}}\\
     \cline{1-7}\noalign{\smallskip}
     Loss&1-shot&5-shot&1-shot&5-shot&1-shot&5-shot\\
     \noalign{\smallskip}
     $\mathcal{L}^{\text{ce}}$&60.2&79.2&68.0&84.7&68.7&86.0\\
     \noalign{\smallskip}
     $\mathcal{L}^{\textrm{ce}}$+$\mathcal{L}^{\textrm{re}}$&48.5&73.2&56.6&82.2&57.1&83.9\\
     \noalign{\smallskip}
     $\mathcal{L}^{\text{ce}}$+$\mathcal{L}^{\text{tp}}$&68.0&81.9&75.4&87.2&75.8&88.2\\
     \noalign{\smallskip}
     $\mathcal{L}^{\text{ce}}$+$\mathcal{L}^{\text{tp}}$+$\mathcal{L}^{\text{re}}$&\textbf{76.9}&\textbf{85.2}&\textbf{82.9}&\textbf{89.1}&\textbf{85.6}&\textbf{91.0}\\
    \hline\noalign{\smallskip}
         \end{tabular}
    \label{tab:ablation}
\end{table}

\begin{figure}[H]
            \centering
            \includegraphics[scale=0.17]{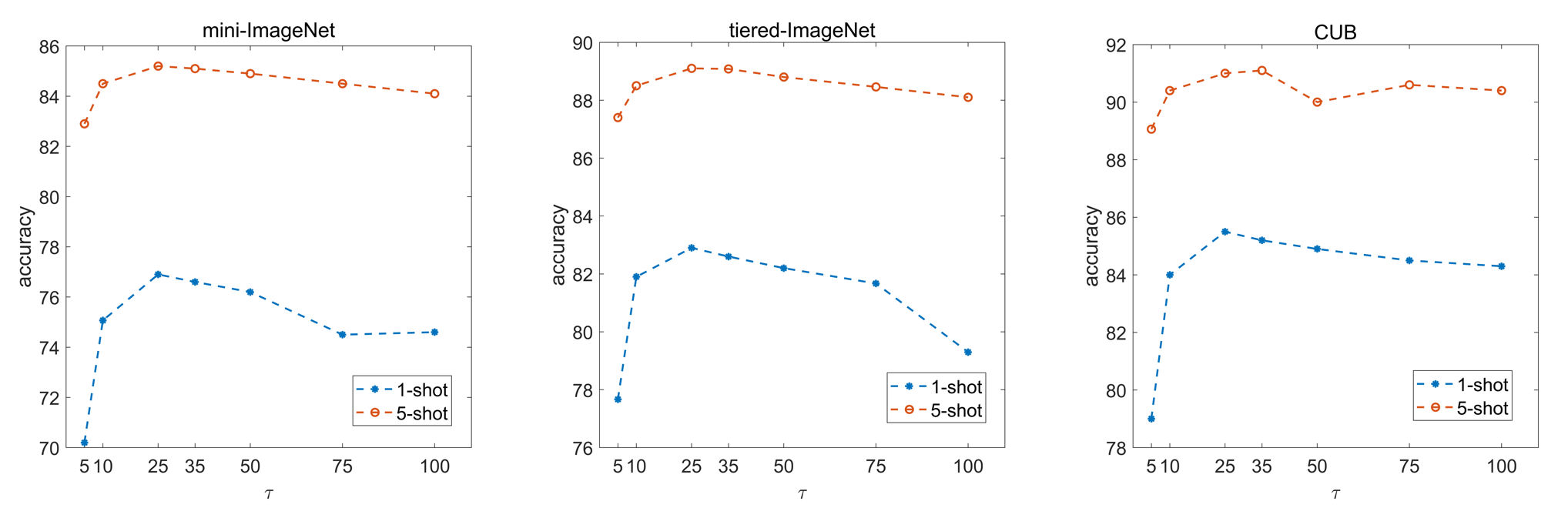}
            \caption{The results for the ablation study of choosing $\tau$  on \textit{mini}-ImageNet, \textit{tiered}-ImageNet, and CUB-200-2011 with ResNet-18  }
            \label{fig:ablation}
        \end{figure}

\noindent\textbf{Choosing the Value of $\tau$:}
The value of $\tau$ denotes the concentration level of the distribution around a centroid. We choose the value of $\tau$ in [5, 10, 25, 35, 50, 75, 100] by ablation study with ResNet-18. The results are reported in Fig. \ref{fig:ablation}. The trends of the results are consistent in both 1-shot and 5-shot scenarios. In most cases, $\tau$= 25 provides the best results in both 1-shot and 5-shot scenarios. So, we set the value of $\tau$ to 25 in all our experiments.

\section{Conclusions}

We proposed a VAE model conditioned on support shots and relaxed a sample level KL regularization constrain to a task level. Our method obtains the state-of-the-art on a wide range of standard few-shot benchmarks. Especially in the one-shot scenario, our method achieves about 3\% improvement compared with previous methods. Besides, we proposed a more challenging scenario for transductive few-shot learning: the query shots of each class are nonuniform. We simulated two nonuniform few-shot scenarios and found that some previous methods suffered performance in this more challenging situation. In comparison, our TP-VAE maintained high performance. Our method has another advantage: TP-VAE is a meta testing module that can connect any base-training module and is a template for researchers to use. In future work, we aim to combine the contrastive learning paradigm, Variational Auto-Encoder paradigm, and few-shot learning paradigm to find a stronger feature extractor and obtain better results.

% \clearpage\mbox{}Page \thepage\ of the manuscript.
% \clearpage\mbox{}Page \thepage\ of the manuscript.

% This is the last page of the manuscript.
% Now we have reached the maximum size of the ECCV 2022 submission (excluding references).
% References should start immediately after the main text, but can continue on p.15 if needed.

\clearpage
\bibliographystyle{splncs04}
\bibliography{eccv2022submission}

\end{document}